\documentclass[Afour,sageh,times]{sagej}
\usepackage{multirow}
\usepackage{moreverb,url}
\usepackage[normalem]{ulem}
\usepackage{ragged2e}
\usepackage{float}

\usepackage[colorlinks,bookmarksopen,bookmarksnumbered,citecolor=red,urlcolor=orange]{hyperref}

\newcommand\BibTeX{{\rmfamily B\kern-.05em \textsc{i\kern-.025em b}\kern-.08em
T\kern-.1667em\lower.7ex\hbox{E}\kern-.125emX}}

\usepackage{amsmath,amssymb,stmaryrd,mathtools}
\usepackage{xcolor}
\usepackage{xspace}
\usepackage{bbm}

\definecolor{orange(sae/ece)}{rgb}{1.0, 0.49, 0.0}
\definecolor{teal(sae/ece)}{rgb}{0, 0.47, 0.52}
\definecolor{purple}{rgb}{0.74, 0.65, 1.0}
\definecolor{dark_purple}{rgb}{0.58, 0.0, 0.82}
\definecolor{light_gray}{rgb}{0.9, 0.9, 0.9}
\definecolor{medium_gray}{rgb}{0.6, 0.6, 0.6} 
\definecolor{dark_gray}{rgb}{0.2, 0.2, 0.2} 
\definecolor{dark_blue}{rgb}{0.098, 0.239, 0.52}
\definecolor{dark_brown}{rgb}{0.3255, 0.004, 0.001}
\definecolor{r3mcolor}{rgb}{0.478, 0.1569, 0.4863}
\definecolor{light_blue}{rgb}{0.33, 0.80, 1}

\newcommand{\rb}[1]{{\color{black} #1}}

\newcommand{\addspace}{\vspace{+0.1cm}}

\newtheorem{definition}{Definition}

\newcommand{\rapl}{\textcolor{orange}{\textbf{RAPL}}\xspace}
\newcommand{\rlhf}{\textcolor{purple}{\textbf{RLHF}}\xspace}
\newcommand{\rlhfl}{\textcolor{dark_purple}{\textbf{RLHF-L}}\xspace}
\newcommand{\tcc}{\textcolor{gray}{\textbf{TCC}}\xspace}
\newcommand{\tccot}{\textcolor{dark_gray}{\textbf{TCC-OT}}\xspace}
\newcommand{\mvpot}{\textcolor{dark_blue}{\textbf{MVP-OT}}\xspace}
\newcommand{\mvpotfinetune}{\textcolor{light_blue}{\textbf{Fine-Tuned-MVP-OT}}\xspace}

\newcommand{\mvpotfinetuneabbrev}{\textcolor{light_blue}{\textbf{FT-MVP-OT}}\xspace}
\newcommand{\imagenetabbrev}{\textcolor{dark_brown}{\textbf{ImNet-OT}}\xspace}

\newcommand{\gt}{\textcolor{teal}{\textbf{GT}}\xspace}
\newcommand{\imagenet}{\textcolor{dark_brown}{\textbf{ImageNet-OT}}\xspace}
\newcommand{\rtm}{\textcolor{r3mcolor}{\textbf{R3M-OT}}\xspace}

\newcommand{\robot}{\mathrm{R}}
\newcommand{\human}{\mathrm{H}}
\newcommand{\repr}{\phi_\robot}
\newcommand{\reph}{\phi_\human}
\newcommand{\approxreph}{\tilde{\phi}_\human}
\newcommand{\rewr}{r} 
\newcommand{\rewh}{r^*} 
\newcommand{\obs}{o}
\newcommand{\obstraj}{\mathbf{o}}

\newcommand{\policyr}{\pi_\robot} 
\newcommand{\policyh}{\pi_\human} 

\newcommand{\actionsR}{\mathbf{a}_\mathrm{R}}%

\usepackage{amsmath}

\DeclareMathOperator*{\argmin}{arg\,min}
\usepackage{colortbl}
\usepackage{multirow}

\begin{document}


\title{Maximizing Alignment with Minimal 
Feedback: Efficiently Learning Rewards for Visuomotor Robot Policy Alignment}

\author{Ran Tian\affilnum{1}, Yilin Wu\affilnum{2}, Chenfeng Xu\affilnum{1},  Masayoshi Tomizuka\affilnum{1},  Jitendra Malik\affilnum{1}, Andrea Bajcsy\affilnum{2}}

\affiliation{\affilnum{1}UC Berkeley\\
\affilnum{2}Carnegie Mellon University\\
This work has been supported in part by the Google Research Scholar Award.}



\begin{abstract}
Visuomotor robot policies, increasingly pre-trained on large-scale datasets, promise significant advancements across robotics domains. 
However, aligning these policies with end-user preferences remains a challenge, particularly when the preferences are hard to specify. 
While reinforcement learning from human feedback (RLHF) has become the predominant mechanism for alignment in non-embodied domains like large language models, it has not seen the same success in aligning visuomotor policies due to the prohibitive amount of human feedback required to learn visual reward functions. 
To address this limitation, we propose Representation-Aligned Preference-based Learning (RAPL), an observation-only method for learning visual rewards from significantly less human preference feedback. Unlike traditional RLHF, RAPL focuses human feedback on fine-tuning pre-trained vision encoders to align with the end-user's visual representation and then constructs a dense visual reward via feature matching in this aligned representation space. 
We first validate RAPL through simulation experiments in the X-Magical benchmark and Franka Panda robotic manipulation, demonstrating that it can learn rewards aligned with human preferences, more efficiently uses preference data, and generalizes across robot embodiments. 
Finally, our hardware experiments align pre-trained Diffusion Policies for three object manipulation tasks. We find that RAPL can fine-tune these policies with 5x less real human preference data, taking the first step towards minimizing human feedback while maximizing visuomotor robot policy alignment. More details (e.g., videos) are at the project \href{https://sites.google.com/berkeley.edu/rapl}{website}.
\end{abstract}

\keywords{Reinforcement learning from human feedback, visuomotor policy learning, representation learning, alignment}


\maketitle

\section{Introduction}

\begin{figure*}[h!]
    \centering
    \includegraphics[width=\textwidth]{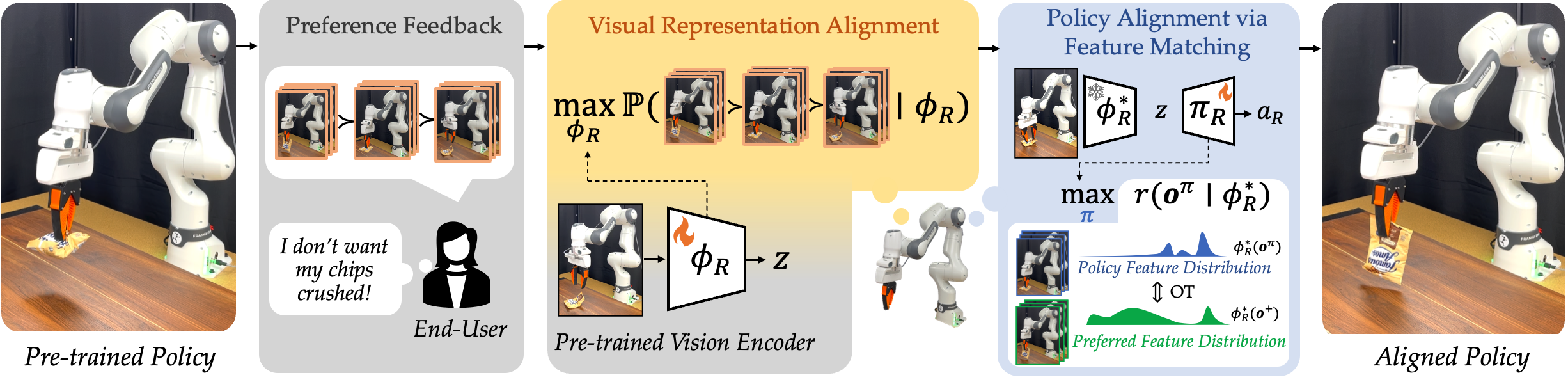}
    \caption{\textbf{Representation-Aligned Preference-based Learning (RAPL)}, is an observation-only method for learning visual robot rewards from significantly less human preference feedback.
    (center) Unlike traditional reinforcement learning from human feedback, RAPL focuses human feedback on fine-tuning pre-trained vision encoders to align with the end-user’s visual representation.
    The aligned representation is used to construct an optimal transport-based visual reward for aligning the robot's visuomotor policy. (left) Before alignment, the robot frequently picks up a bag of chips by squeezing the middle, risking damage to the contents. (right) After alignment with our RAPL reward, the robot adheres to the end-user's preference and picks up the bag by its edges.}
    \label{fig:front_fig}
\end{figure*}

Visuomotor robot policies---which predict actions directly from high-dimensional image observations---are being revolutionized by pre-training on large-scale datasets. 
For example, robots like manipulators \citep{brohan2023rt, chi2024diffusionpolicy}, humanoids \citep{radosavovic2023real}, and autonomous cars \citep{hu2023planning, tiantokenize} rely on pre-trained vision encoders \citep{deng2009imagenet} for representing RGB images and for learning multimodal behavior policies from increasingly large observation-action teleoperation datasets \citep{padalkar2023open}. 

Despite this remarkable progress, these visuomotor policies do not always act in accordance with human end-user preferences. 
For instance, consider the scenario on the left of \autoref{fig:front_fig} where a robot manipulator is trained to imitate diverse teleoperators picking up a bag of chips. At deployment, the end-user prefers the chips to remain intact. However, the manipulator frequently grasps the bag by squeezing the middle—risking damage to the chips—instead of holding the packaging by its edges like the user prefers.

A foundational approach to tackle this \textit{misalignment} between a pre-trained policy and an end-user's hard-to-specify preferences is reinforcement learning from human feedback (RLHF) \citep{christiano2017deep}. 
By presenting the end-user with outputs generated by the pre-trained model and collecting their preference rankings, RLHF trains a reward model that is then used to fine-tune the base policy, enabling it to produce outputs that better align with the end-user's preferences. 
While RLHF has emerged as the predominant alignment mechanism in non-embodied domains such as large language models (LLMs) \citep{ouyang2022training} and text-to-image generation models \citep{lee2023aligning}, it has not shown the same impact for aligning visuomotor robot policies. 
Fundamentally, this challenge arises because learning a high-quality visual reward function requires an impractically large amount of human preference feedback: in our hardware experiments, collecting \textbf{200 preference rankings} for a \textbf{single task} takes approximately \textbf{1 day}. 
If we hope to align generalist, pre-trained visuomotor policies, it is essential to adapt the RLHF paradigm to operate with significantly less human feedback.  

Our approach is motivated by seminal work in inverse reinforcement learning \citep{abbeel2004apprenticeship, ziebart2008maximum}, which states that the desired reward function we seek to learn via RLHF makes the robot's behavior \textit{indistinguishable} from the human's ideal behavior. 
Mathematically, 
this reward can be modeled as the divergence between the feature distribution of the robot’s policy and that of the end-user’s optimal demonstrations \citep{pomerleau1988alvinn, dadashi2020primal, sun2019provably, swamy2021moments}.
However, when constructing visual rewards, this feature matching has to occur within the \textit{end-user’s visual representation of the world}, which is unknown to the robot a priori.  
This implies that a critical aspect of reward learning is identifying which visual features are important to the end-user.
Instead of jointly learning visual features and a divergence measure through human feedback \citep{christiano2017deep}, our key idea is to allocate the limited human preference budget \textit{exclusively} to fine-tuning pre-trained vision encoders, aligning their visual representations with those of the end-user.
Once the visual representation is fine-tuned, the reward function can be directly instantiated as dense feature matching using techniques such as optimal transport \citep{kantorovich1958space} within this aligned visual representation space (center, \autoref{fig:front_fig}).  

In Section~\ref{sec:RAPL method}, we formalize the visual representation alignment problem for robotics as a metric learning problem in the human's representation space. 
We then propose \textbf{R}epresentation-\textbf{A}ligned \textbf{P}reference-based \textbf{L}earning (\textbf{RAPL}), a tractable observation-only method for learning visual robot rewards for aligning visuomotor robot policies. 
%
In Section~\ref{sec:RAPL-result}, 
%
we perform simulation experiments in the X-Magical benchmark and in manipulation with the Franka Panda robot, ablating the way we learn visual representations and how we design reward predictors.
We find that RAPL learns visual rewards that closely align with ground truth rewards,  
enables a more human data-efficient RLHF paradigm for aligning robot visuomotor policies, 
and shows strong zero-shot generalization when the visual representation is learned on a different embodiment from the robot's.
Finally, in Section~\ref{sec:DPO-result}, we instantiate RAPL in hardware experiments for efficiently aligning Diffusion Policies \citep{chi2024diffusionpolicy} deployed in three real world object manipulation tasks: chip bag grasping, cup picking, and fork pick-and-place into a bowl. 
We instantiate RAPL in the context of direct preference optimization \citep{rafailov2024direct}, a variant of RLHF that aligns generative models directly using preference rankings without requiring reinforcement learning. 
RAPL's visual reward reduces reliance on real human preference rankings by \textbf{5x}, generating high-quality synthetic preference rankings at scale and aligning the robot's visuomotor policy with the human's preferences (right, \autoref{fig:front_fig}).

This work is an extended version of the conference paper \cite{tian2023matters}. We expand the content of this paper via:
\begin{itemize}
    \item \textbf{New ablations and baselines in simulation}. In Section~\ref{subsec:in-domain-manipulation}, we expanded our evaluation by including additional visual reward baselines, providing visualizations of the visual representations, and evaluating RAPL on a cluttered robot manipulation task with visual distractors.
    These further validate RAPL’s ability to disentangle the visual features that influence end-user preferences.
    \vspace{0.5em}
    \item \textbf{Aligning generative robot policies in the real world.} In Section~\ref{sec:DPO-result}, we extended our approach from simulation to hardware. We use RAPL as a synthetic preference generator built from only 20 real human preference rankings and use the rankings to align a diffusion-based visuomotor policy in three real-world manipulation tasks. 
    These results highlight how RAPL can  alleviate human feedback burden while still ensuring high-quality policy alignment. 
\end{itemize}
\section{Related Work}
\label{sec:related_work}

\textbf{Preference alignment of generative models.} Generative models, such as large language models (LLMs), text-to-image generation models, and behavior cloning models, are predominantly trained using an imitative objective. While this approach simplifies training with internet-scale data, this objective serves only as a ``proxy" for the true training goal: the human internal reward function. As a result, these generative models may be misaligned with end-user preferences or may even lead to safety-critical scenarios (e.g., generating unsafe texts \citep{murule}, images \citep{lee2023aligning}, and robot motions \citep{lu2023imitation}).
Preference alignment, particularly through reinforcement learning from human feedback (RLHF) \citep{christiano2017deep}, has emerged as a key strategy for aligning generative models with human preferences, especially in non-embodied domains.
RLHF involves three key stages: feedback elicitation (e.g., presenting users with two or more model generations and gathering preference rankings), reward modeling (e.g., training a reward model to replicate the rankings), and policy optimization via reinforcement learning (RL) \citep{schulman2017proximal}. Traditional RL methods require a simulator to gather the learning agent's experiences under the current reward model. 
To address this dependency, direct preference optimization (DPO) \citep{rafailov2024direct} has been proposed as a variant of RLHF that directly updates the generative model using preference rankings through contrastive learning.
Despite the success of RLHF in non-embodied domains, and regardless of its algorithmic instantiation, its application in embodied settings remains prohibitively expensive due to the challenges of collecting extensive human feedback on robot rollouts.
Our work seeks to adapt the RLHF paradigm for embodied domains, enabling efficient visuomotor robot policy alignment. 

\noindent
\textbf{Scaling preference alignment with synthetic feedback.}
Recent works in the LLM domain have explored leveraging AI feedback to automatically generate preference rankings, thereby facilitating the reward learning. These approaches typically rely on either a single teacher model \citep{bai2022constitutional, leerlaif, murule}, an ensemble of teacher models \citep{tunstall2023zephyr}, or targeted context prompting techniques, which prompt the model to generate both positive and negative outputs for constructing comparison-based preference rankings \citep{yang2023rlcd, liu2024direct}.
However, these methods are not directly applicable to embodied contexts, such as robotic manipulation, due to the lack of high-quality, open-source, input-unified foundation models. 
Input-unified foundation models are particularly challenging in robotics because different embodied models often rely on incompatible input modalities or features, making it difficult to transfer feedback effectively across models.
Unlike previous works that focus on scaling human feedback, our approach aims to effectively learn end-user rewards from a limited amount of human feedback.

\noindent
\textbf{Visual reward learning in robotics.}
Visual reward models aim to capture task preferences directly from image observations. 
Self-supervised approaches leverage task progress inherent in video demonstrations to learn how ``far'' the robot is from completing the task \citep{zakka2022xirl, kumar2023graph, ma2023vip} 
while other approaches identify task segments and measure distance to these subgoals \citep{sermanet2016unsupervised, tanwani2020motion2vec, shao2020concept, chen2021learning}. 
However, these approaches fail to model preferences \textit{during} task execution that go beyond progress (e.g., spatial regions to avoid during movement). 
Fundamental work in IRL uses feature matching between the expert and the learner in terms of the expected state visitation distribution to infer rewards \citep{abbeel2004apprenticeship, ziebart2008maximum}, and recent work in optimal transport (OT) has shown how to scale this matching to high dimensional state spaces \citep{xiao2019wasserstein, dadashi2020primal, papagiannis2022imitation, luo2023optimal}. 
However, the key to making this matching work from high-dimensional visual input spaces is a good visual embedding. 
Recently there has been excitement in using representations from pre-trained visual models, but key to making these work in robotics is fine-tuning, which is typically done via proxy tasks like dynamics prediction or enforcing temporal cycle-consistency \citep{dadashi2020primal}. However, all these proxy tasks bypass the human’s input on what matters to them, exacerbating spurious correlations and ultimately leading to robot behaviors that are misaligned with user preferences.
In contrast to prior works that rely on using only self-supervised signals, we propose an OT-based visual reward that is trained purely on videos (no action labels needed) that are ranked by the \textit{end-user's} preferences. 

\addspace
\noindent
\textbf{Representation alignment in robot learning.} Representation alignment studies the agreement between the representations of two learning agents.
As robots will ultimately operate in service of people, representation alignment is becoming increasingly important for robots to interpret the world in the same way as we do.
Previous work has leveraged user feedback, such as human-driven feature selection \citep{bullard2018human, luu2019incremental}, interactive feature construction \citep{bobu2021feature, katz2021preference}, or similarity-implicit representation learning \citep{bobu2023sirl}, to learn aligned representations for robot behavior learning. But they either operate on a manually defined feature set or learning features in low-dimensional state space settings (e.g., positions).
In the visual domain,
\citep{zhang2020learning} uses a per-image reward signal to align the image representation with the preferences encoded in the reward signal; however, when the main objective is learning the human's reward then assuming a priori access to such a reward signal is not feasible. 
Instead, our work utilizes human preference feedback to align the robot's visual representations with the end user. 

\section{Problem Setup}
\label{sec:problem-setup}

\textbf{Human policy.} 
We consider scenarios where the robot R wants to learn how to perform a task for human H. 
The human knows the desired reward $\rewh$ which encodes their preferences for the task. The human acts via an approximately optimal policy $\policyh^{*}: \mathcal{O} \rightarrow \Delta^{n_{\mathcal{A}_\human}}$ optimized under their underlying reward function. Here, $\Delta$
denotes the probability simplex over the human’s $n_{\mathcal{A}_\human}$-dimensional action space.
Instead of directly consuming the raw perceptual input, research suggests that humans naturally build visual representations of the world \citep{bonnen2021ventral} that focus on task-relevant attributes \citep{callaway2021fixation}. 
We model the human's representation model as $\reph: \mathcal{O} \rightarrow Z_{\mathrm{H}}$, mapping from the perceptual input $o \in \mathcal{O}$ to the human's latent space $Z_{\mathrm{H}}$ which captures their task and preference-relevant features.

\noindent
\textbf{Visumotor policy alignment}. We denote the robot's visumotor policy as $\policyr: \mathcal{O} \rightarrow \Delta^{n_{\mathcal{A}_\robot}}$ where $\Delta$ denotes the probability simplex over the robot's $n_{\mathcal{A}_\robot}$-dimensional action space. 
In the RLHF paradigm, we seek to fine-tune $\policyr$ to maximize a reward function: 
\begin{equation}
\begin{aligned}
\policyr^* = \arg\max_{\policyr}  \mathbb{E}_{\obstraj \sim p(\obstraj \mid \policyr)} \Big[ \sum_{t=0}^{\infty} \gamma^t \cdot \rewr(\repr(\obs^t))\Big],
\end{aligned}
\label{eq:robot_planning_problem}
\end{equation}
where $\gamma \in [0,1)$ is the discount factor and $\obstraj = \{\obs^0, \obs^1, \hdots \}$ is the image observation trajectory induced by the robot's policy. 
The robot's reward $\rewr$ also relies on a visual representation, $\repr: \mathcal{O} \rightarrow Z_\robot$, which maps from the image observation $\obs \in \mathcal{O}$ to a lower-dimensional latent space, $Z_\robot$. In general, this could be hand-crafted, such as distances to objects from the agent's end-effector \citep{ziebart2008maximum, levine2011nonlinear, finn2016guided}, or the output of an encoder pre-trained on large-scale datasets \citep{chen2021learning, ma2023vip}. The optimization problem in \autoref{eq:robot_planning_problem} can be approximately solved through reinforcement learning algorithm such as PPO \citep{schulman2017proximal}, where the robot repeatedly interacts with a simulated environment and receives reward feedback to improve its policy. 
When a simulated environment is not accessible or real-world interaction is prohibitively expensive, direct preference optimization (DPO) \citep{rafailov2024direct} has been proposed as a variant of RLHF that directly updates the generative model through contrastive learning using preference rankings drawn from reward $\rewr$ \citep{liu2024direct}:
\begin{flalign}
\max_{\policyr} \mathbb{E}_{(\actionsR^{+}, \actionsR^{-}, o)\sim \mathcal{D}_{\text{pref}}} \mathbb{P} [\actionsR^{+} \succ \actionsR^{-} | o],  \label{eq:dpo_update} 
\end{flalign}
\begin{flalign}
\mathbb{P} [\actionsR^{+} \succ \actionsR^{-} | o]  =
   -\log \frac{\exp \big(\alpha \log \frac{\policyr(\actionsR^+, o)}{\pi_{\text{ref}}(\actionsR^-, o)}\big)}{\sum\limits_{\actionsR \in \{\actionsR^+, \actionsR^-\}} \exp \big(\alpha \log \frac{\policyr(\actionsR, o)}{\pi_{\text{ref}}(\actionsR, o)}\big) } , \nonumber 
\end{flalign}
\noindent
where $\pi_{\text{ref}}$ is the robot's initial reference visumotor policy, $\mathcal{D}_{\text{pref}}$ is a preference dataset. 
Each training example consists of $(\actionsR^+, \actionsR^-, o)$: an initial observation $o$ and a pair of action sequences, $\actionsR^+$ and $\actionsR^-$, sampled from the reference policy given this initial observation. The plus and minus denote preferred and unpreferred behaviors under the human's internal reward such that 
$\actionsR^{+} \succ \actionsR^{-} | o \implies \sum_{t}r(o^+_t) > \sum_{t}r(o^-_t)$, where $o^{+}_t, o^{-}_t$ denotes the observations when rolling out $\actionsR^+$ and $\actionsR^-$ respectively.
Before the robot can optimize for $\policyr$ using either approach, it is faced with two questions: what visual representation $\repr$ should it use to encode the environment, and which reward $\rewr$ should it optimize to align its behavior with $\policyh^{*}$?

\section{RAPL: Representation-Aligned Preference-Based Learning}
\label{sec:RAPL method}

\subsection{The Visual Representation Alignment Problem for Robotics}
\label{subsec:rapl_formalism}


We follow the formulation in \cite{sucholutsky2023alignment} and bring this to the robot learning domain. Intuitively, visual representation alignment is defined as the degree to which the output of the robot's encoder, $\repr$, matches the human's internal representation, $\reph$, for the same image observation, $
\obs \in \mathcal{O}$, during task execution.
We utilize a triplet-based definition of representation alignment as in \citep{ jamieson2011low} and \citep{sucholutsky2023alignment}.

    
\begin{definition}[\textbf{Triplet-based Representation Space}] Let $\obstraj = \{\obs^t\}^T_{t=0}$ be a sequence of image observations over $T$ timesteps, $\phi : \mathcal{O} \rightarrow Z$ be a given representation model, and $\phi_{}(\obstraj) := \{\phi(\obs^{0}),\dots,\phi(\obs^{T})\}$ be the corresponding embedding trajectory. For some distance metric $d(\cdot, \cdot)$ and two observation trajectories $\obstraj^{i}$ and $\obstraj^{j}$, let $d\big(\phi_{}(\obstraj^{i}),\phi_{}(\obstraj^{j})\big)$ be the distance between their embedding trajectories. The triplet-based representation space of $\phi$ is:
\begin{equation}
\begin{aligned}
S_{\phi_{}} = \Big\{(\obstraj^{i},\obstraj^{j},\obstraj^{k}) : \obstraj^{j|i} \succ \obstraj^{k|i}, \obstraj^{i,j,k} \in \Xi \Big\},\label{eq:representation_space}
\end{aligned}
\end{equation}
where $\Xi$ is the set of all possible image trajectories for the task of interest, and $\obstraj^{j|i} \succ \obstraj^{k|i}$ denotes $d\big(\phi_{}(\obstraj^{i}),\phi_{}(\obstraj^{j})\big) < d\big(\phi_{}(\obstraj^{i}),\phi_{}(\obstraj^{k})\big)$.
\end{definition}



Intuitively, this states that the visual representation $\phi$ helps the agent determine how similar two videos are in a lower-dimensional space. For all possible triplets of videos that the agent could see, it can determine which videos are more similar and which videos are less similar using its embedding space. The set $S_{\phi}$ contains all such similarity triplets. 


\begin{definition}[\textbf{Visual Representation Alignment Problem}]
Recall that $\reph$ and $\repr$ are the human and robot's visual representations respectively. The representation alignment problem is defined as learning a $\repr$ which minimizes the difference between the two agents' representation spaces, as measured by a function $\ell$ which penalizes divergence between the two representation spaces:
\begin{equation}
\begin{aligned}
\min_{\phi_{\mathrm{R}}} \ell(S_{\phi_{\mathrm{R}}}, S_{\phi_{\mathrm{H}}}).
\end{aligned}
\label{eq:representation_alignment_problem}
\end{equation}
\end{definition}

\subsection{Representation Alignment via Preference-based Learning}
\label{subsec:rapl_inference}

Although this formulation sheds light on the underlying problem, solving \autoref{eq:representation_alignment_problem} exactly is impossible since the functional form of the human's representation $\phi_{\mathrm{H}}$ is unavailable and the set $S_{\phi_{\mathrm{H}}}$ is infinite. Thus, we approximate the problem by constructing  a subset $\tilde{S}_{\reph} \subset S_{\reph}$ of triplet queries. 
Since we seek a representation that is relevant to the human's preferences, we ask the human to rank these triplets based on their \textit{preference-based} notion of similarity (e.g., $\rewh(\reph(\obstraj^i)) > \rewh(\reph(\obstraj^j)) > \rewh(\reph(\obstraj^k)) \implies \obstraj^{j|i} \succ \obstraj^{k|i}$).
With these rankings, we implicitly learn $\reph$ via a neural network trained on these triplets.


We interpret a human’s preference over the triplet $(\obstraj^{i},\obstraj^{j},\obstraj^{k}) \in \tilde{S}_{\reph}$ via the Bradley-Terry model \citep{bradley1952rank}, where $\obstraj^{i}$ is treated as an anchor and $\obstraj^{j}, \obstraj^{k}$ are compared to the anchor in terms of preference similarity as in \autoref{eq:representation_space}:
\begin{flalign}
&\mathbb{P}(\obstraj^{j|i} \succ \obstraj^{k|i} \mid \reph) \nonumber \\
&\approx \frac{e^{-d(\reph(\obstraj^i), ~\reph(\obstraj^j))}}{e^{-d(\reph(\obstraj^i), ~\reph(\obstraj^j))} + e^{-d(\reph(\obstraj^i), ~\reph(\obstraj^k))}}. \label{eq:representation_alignment_loss}
\end{flalign}


One remaining question is: what distance measure $d$ should we use to quantify the difference between two embedding trajectories? In this work, we use optimal transport as a principled way to measure the feature matching between any two videos. 
For any video $\obstraj$ and for a given representation $\phi$, let the induced empirical embedding distribution be $\rho = \frac{1}{T} \sum_{t=0}^{T} \delta_{\phi(\obs^t)}$, where $\delta_{\phi(\obs^t)}$ is a Dirac distribution centered on $\phi(\obs^t)$. 
Optimal transport finds the optimal transport plan $\mu^* \in \mathbb{R}^{T \times T}$ that transports one embedding distribution, $\rho_i$, to another video embedding distribution, $\rho_j$, with minimal cost. 
This comes down to an optimization problem that minimizes the Wasserstein distance between the two distributions: 
\begin{equation}
\begin{aligned}
\mu^{*} = \argmin_{\mu \in \mathcal{M}(\rho_i, \rho_j)} \sum_{t=1}^{T}\sum_{t'=1}^{T} c\big(\phi(\obs^t_i), \phi(\obs^{t'}_j)\big)\mu_{t,t'}.
\end{aligned}
\label{eq:app_ot_problem}
\end{equation}
where $\mathcal{M}(\rho_i, \rho_j) = \{\mu \in \mathbb{R}^{T \times T}: \mu\mathbf{1} = \rho_{i}, \mu^{T}\mathbf{1} = \rho_j\}$ is the set of transport plan matrices, $c: \mathbb{R}^{n_e} \times \mathbb{R}^{n_e}\rightarrow \mathbb{R}$ is a cost function defined in the embedding space (e.g., cosine distance), and $n_e$ is the dimension of the embedding space.
Solving the above optimization in \autoref{eq:app_ot_problem} exactly is generally intractable for high dimensional distributions. In practice, we solve an entropy-regularized version of the problem following the Sinkhorn algorithm \citep{peyre2019computational} which is 
amenable to fast optimization:
\begin{equation}
    \begin{aligned}
    \hspace{-0.1cm}\mu^{*} = \hspace{-0.1cm}\argmin_{\mu \in \mathcal{M}(\rho_i, \rho_j)} \sum_{t=1}^{T}\sum_{t'=1}^{T} c\big(\phi(\obs^t_i), \phi(\obs^t_j)\big)\mu_{t,t'} - \epsilon\mathcal{H}(\mu),
    \end{aligned}
\label{eq:app_ot_problem_regu}
\end{equation}
where $\mathcal{H}$ denotes the entropy term that regularizes the optimization and $\epsilon$ is the associated weight. 

The optimal transport plan gives rise to the following distance that measures the feature matching between any two videos and is used in \autoref{eq:representation_alignment_loss}: 
\begin{equation}
d(\phi(\obstraj^i), ~\phi(\obstraj^j)) = -\sum_{t=1}^{T}\sum_{t'=1}^{T} c\big(\phi(\obs^{t}_i), \phi(\obs^{t'}_j)\big)\mu^{*}_{t, t'}.
\end{equation}

Our final optimization is a maximum likelihood estimation problem:
\begin{align}
    \approxreph := \max_{\reph} \sum_{(\obstraj^{i},\obstraj^{j},\obstraj^{k})\in\tilde{S}_{\reph}} \mathbb{P}(\obstraj^{j|i} \succ \obstraj^{k|i} \mid \reph).
    \label{eq:representation_optimization}
\end{align}
Since the robot seeks a visual representation that is aligned with the human's, we set: 
\begin{equation}
    \repr := \approxreph.
\end{equation}

\subsection{Preference-aligned Visual Reward Model}
\label{subsec:rapl_visual_reward}

Given our aligned visual representation, we seek a robot visual reward function $r$ that approximates the end-user's reward function for aligning the robot's visuomotor policy with human preference. Traditional IRL methods \citep{abbeel2004apprenticeship, ziebart2008maximum} are built upon matching the feature distribution of the robot’s policy with that of the end-user’s demonstration. Specifically, we seek to match the observation distribution induced by the robot's policy $\policyr$, and the observation distribution of a human's preferred video demonstration, $\obstraj_+$ in the aligbned representation space.

The optimal transport plan between the two distributions precisely defines a reward function that measures this matching \citep{kantorovich1958space} and yields the reward which is optimized in \autoref{eq:robot_planning_problem}:
\begin{equation}
\begin{aligned}
\rewr(\obs^t_\mathrm{R}; \repr, \obstraj_+) = -\sum_{t'=1}^{T} c\big(\repr(\obs^{t}_\robot), \repr(\obs^{t'}_+)\big)\mu^{*}_{t, t'}.
\end{aligned}
\label{eq:ot_reward}
\end{equation}
%
This reward has been successful in prior vision-based robot learning \citep{haldar2023teach, haldar2023watch, guzey2023see} with the key difference in our setting being that we use RAPL's aligned visual representation $\repr$ for feature matching. 

\section{Reinforcement Learning in Simulation with RAPL}
\label{sec:RAPL-result}

In this section, we design a series of experiments to evaluate RAPL’s effectiveness in learning visual rewards and aligning the robot visuomoto policy in simulations through reinforcement learning.
%

\subsection{Experimental Design }

\textbf{Preference dataset: $\tilde{S}_{\reph}$.} While the ultimate test is learning from real end-user feedback, in this section, we first use a simulated human model, which allows us to easily ablate the size of the preference dataset, and gives us privileged access to $\rewh$ for direct comparison.
In all environments, the simulated human constructs the preference dataset $\tilde{S}_{\reph}$ by sampling triplets of videos uniformly at random from the set\footnote{To minimize the bias of this set, we construct $\tilde{\Xi}$ such that the reward distribution of this set under $\rewh$ is approximately uniform. Future work should investigate study this set design further, e.g., \citep{sadigh2017active}.} of video observations $\tilde{\Xi} \subset \Xi$, and then ranking them with their reward $\rewh$ as in  \autoref{eq:representation_alignment_loss}. 

\addspace
\noindent
\textbf{Independent \& dependent measures.} Throughout our experiments, we vary the \textit{visual reward signal} used for robot policy optimization and the \textit{preference dataset size} used for representation learning. 
We measure robot task success as a binary indicator of if the robot completed the task with high reward $\rewh$. 

\addspace
\noindent
\textbf{Controlling for confounds.} 
Our ultimate goal is to have a visual robot policy, $\policyr$, that takes as input observations and outputs actions. 
However, to rigorously compare policies obtained from different visual rewards, we need to disentangle the effect of the reward signal from any other policy design choices, such as the input encoders and architecture. 
To have a fair comparison, we follow the approach from \citep{zakka2022xirl, kumar2023graph} and input the privileged ground-truth state into all policy networks, but vary the visual reward signal used during policy optimization. 
Across all methods, we use an identical reinforcement learning setup and Soft-Actor Critic (SAC) for training \citep{haarnoja2018soft} with code base from \citep{zakka2022xirl}.
When running SAC, the reward (\autoref{eq:ot_reward}) requires matching the robot to an expert demonstration video. For all policy learning experiments, we use $10$ expert demonstrations as the demonstration set $\mathcal{D}_+$ for generating the reward.
To choose this expert observation, we follow the approach from \citep{haldar2023watch}. 
During policy optimization, given a robot's trajectory's observation $\obstraj_\mathrm{R}$ induced by the robot policy $\policyr$, we select the the ``closest" expert demonstration $\obstraj^*_+ \in \mathcal{D}_+$ to match the robot behavior with. This demonstration selection happens via:
\begin{equation}
\begin{aligned}
\hspace{-0.1cm}\obstraj^*_+ \hspace{-0.1cm}= \hspace{-0.1cm}\argmin_{\obstraj+ \in \mathcal{D_+}} \hspace{-0.1cm}\min_{\mu \in \mathcal{M}(\rho_\robot, \rho_+)} \sum_{t=1}^{T}\hspace{-0.1cm}\sum_{t'=1}^{T} \hspace{-0.1cm}c\big(\phi(\obs^t_\robot), \phi(\obs^{t'}_+)\big)\mu_{t,t'}.
\end{aligned}
\label{eq:select_expert_obs}
\end{equation}

\begin{figure}[!t]
    \centering
    \includegraphics[width=0.49\textwidth]{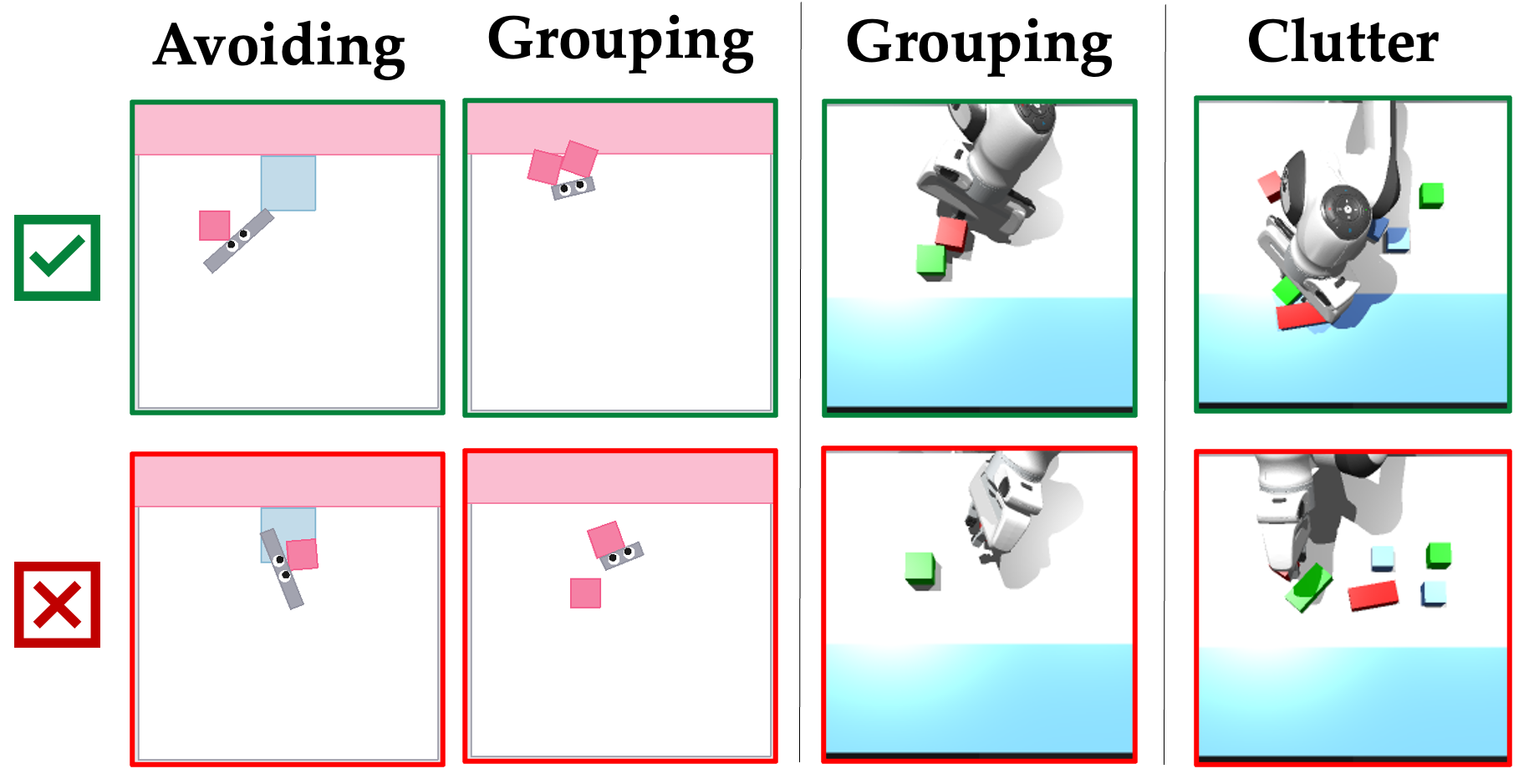}
  \caption{X-Magical \& IsaacGym tasks.
  Top row are high-reward behaviors and bottom row are low-reward behaviors according to the human's preferences. }
  \vspace{-0.5cm}
  \label{fig:x-magical-env}
\end{figure}

\begin{figure*}[t!]
    \centering
    \includegraphics[width=\textwidth]{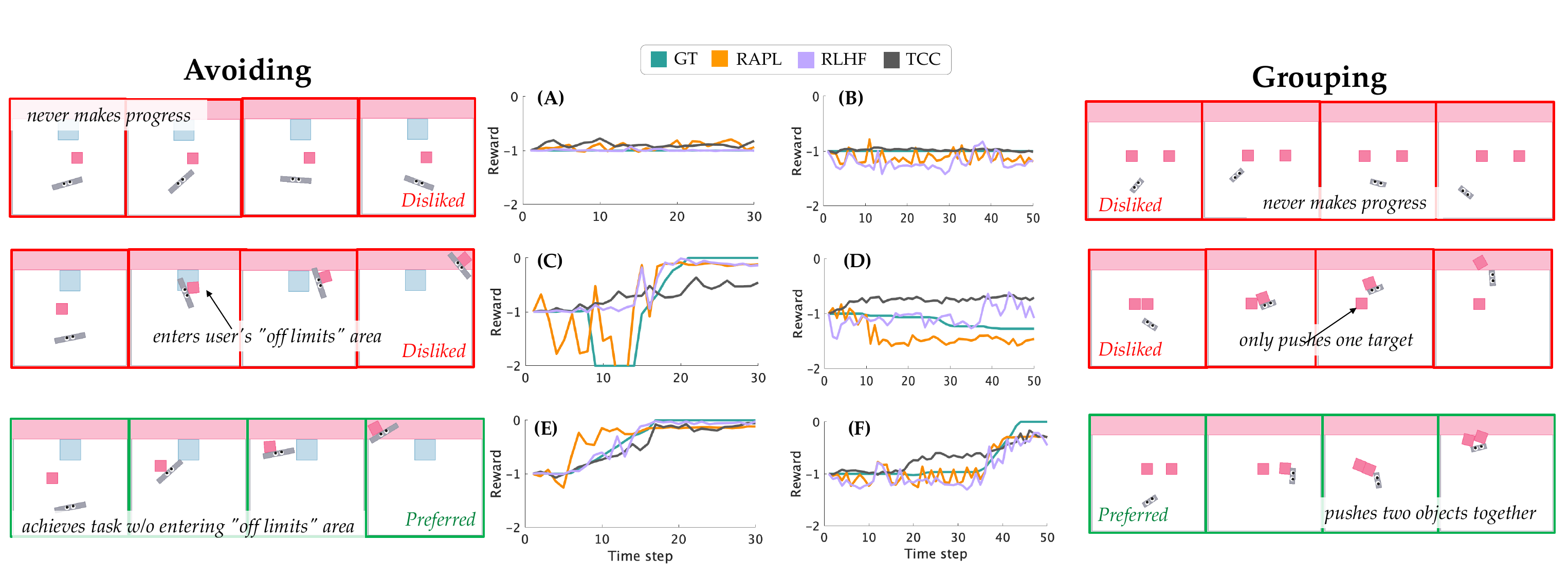}
    \vspace{-1em}
    \caption{\textbf{X-Magical.} (left \& right) examples of preferred and disliked videos for each task. (center) reward associated with each video under each method. RAPL's predicted reward follows the GT pattern: low reward when the behavior are disliked and high reward when the behavior are preferred. RLHF and TCC assign high reward to disliked behavior (e.g., (D)). }
    \vspace{-1em}
    \label{fig:xmagical_indomain_reward}
\end{figure*}

In Section~\ref{sec:in-domain-results}, we first control the agent's embodiment to be consistent between both representation learning and robot optimization (e.g., assume that the robot shows video triplets of itself to the human and the human ranks them). 
In Section~\ref{sec:cross-domain-results}, we relax this assumption and consider the more realistic cross-embodiment scenario where the representation learning is performed on videos of a different embodiment 
than the robot's.



\subsection{Results: From Representation to Policy Alignment}
\label{sec:in-domain-results}

We first experiment in the X-Magical environment \citep{zakka2022xirl}, and then in the realistic IsaacGym simulator. 

\subsubsection{X-Magical Experiments}
\label{subsec:in-domain-xmagical}

\begin{center}
\justifying
\textbf{Tasks.}  We design two tasks inspired by kitchen countertop cleaning. The robot always has to push objects to a goal region (e.g., trash can), shown in pink at the top of the scene in \autoref{fig:x-magical-env}. In the \textbf{avoiding} task, the end-user prefers that the robot and objects never enter an \textit{off-limits zone} during pushing (blue box in left \autoref{fig:x-magical-env}).
In the \textbf{grouping} task, the end-user prefers that objects are pushed \textit{efficiently together} (instead of one-at-a-time) to the goal (center, \autoref{fig:x-magical-env}).
\end{center}

\noindent
\textbf{Privileged state \& reward.}
For \textbf{avoiding}, the true state $s$ is 7D: planar robot position ($p_\robot \in \mathbb{R}^2$) and orientation ($\theta_\mathrm{R}$),  planar position of the object ($p_{\mathrm{obj}}\in \mathbb{R}^2$), distance between goal region and object, ($d_{\mathrm{obj2goal}}$), and distance between the off-limits zone and the object ($d_{\mathrm{obs2obj}}$). The human's reward is: $\rewh_{\mathrm{avoid}}(s) = - d_{\mathrm{goal2obj}} - 2\cdot \mathbb{I}(d_{\mathrm{obs2obj}} < d_{\mathrm{safety}})$, where $d_{\mathrm{safety}}$ is a safety distance and $\mathbb{I}$ is an indicator function giving $1$ when the condition is true.
For \textbf{grouping}, the state is 9D: $s := (p_\robot, \theta_\robot$, $p_{\mathrm{obj^1}}, p_{\mathrm{obj^2}}, d_{\mathrm{goal2obj^1}}$, $d_{\mathrm{goal2obj^2}})$. The human's reward is: $\rewh_{\mathrm{group}}(s) = - \max(d_{\mathrm{goal2obj^1}}, d_{\mathrm{goal2obj^2}}) - ||p_{\mathrm{obj^1}} - p_{\mathrm{obj^2}}||_2$.

\addspace
\noindent
\textbf{Baselines.} We compare our visual reward, \rapl, against (1) \gt, an oracle policy obtained under $\rewh$, (2) \rlhf, which is vanilla preference-based reward learning \citep{christiano2017deep, brown2019extrapolating} that directly maps an image observation to a scalar reward, and (3) \tcc \citep{zakka2022xirl,kumar2023graph} which finetunes a pre-trained encoder via temporal cycle consistency constraints using 500 task demonstrations and then uses L2 distance between the current image embedding and the goal image embedding as reward. We use the same preference dataset with $150$ triplets for training RLHF and RAPL. 

\begin{figure}[t!]
    \centering
    \includegraphics[width=0.35\textwidth]{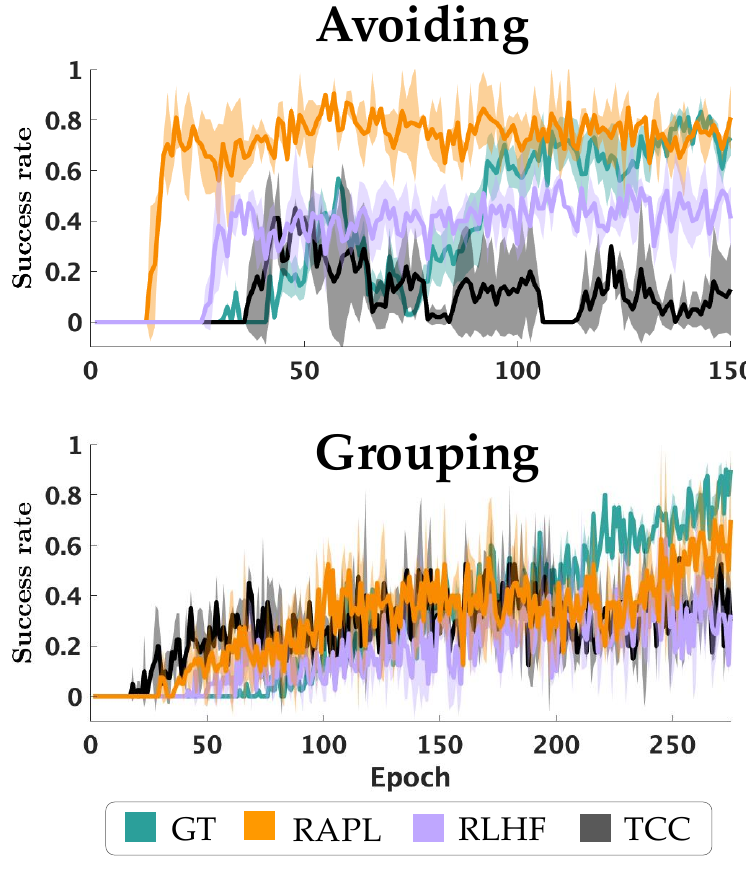}
    \caption{\textbf{X-Magical.} Policy evaluation success rate during policy learning. Colored lines are the mean and variance of the evaluation success rate. RAPL can match GT in the \textbf{avoiding} task and outperforms baseline visual rewards in \textbf{grouping} task.}
    \vspace{-1em}
    \label{fig:xmagical_indomain_policy}
\end{figure}

\addspace
\noindent
\textbf{Visual model backbone.} We use the same setup as in \citep{zakka2022xirl} with the ResNet-18 visual backbone \citep{he2016deep} pre-trained on ImageNet. The original classification head is replaced with a linear layer that outputs a $32$-dimensional vector as our embedding space, $Z_\robot := \mathbb{R}^{32}$. The TCC representation model is trained with 500 demonstrations using the code from \citep{zakka2022xirl}. Both RAPL and RLHF only fine-tune the last linear layer. All representation models are frozen during policy learning.

\addspace
\noindent
\textbf{Hypothesis}. \textit{\rapl is better at capturing preferences \textit{beyond} task progress compared to direct reward prediction \rlhf or \tcc visual reward, yielding higher success rate.} 


\begin{figure*}[t!]
    \centering
    \includegraphics[width=\textwidth]{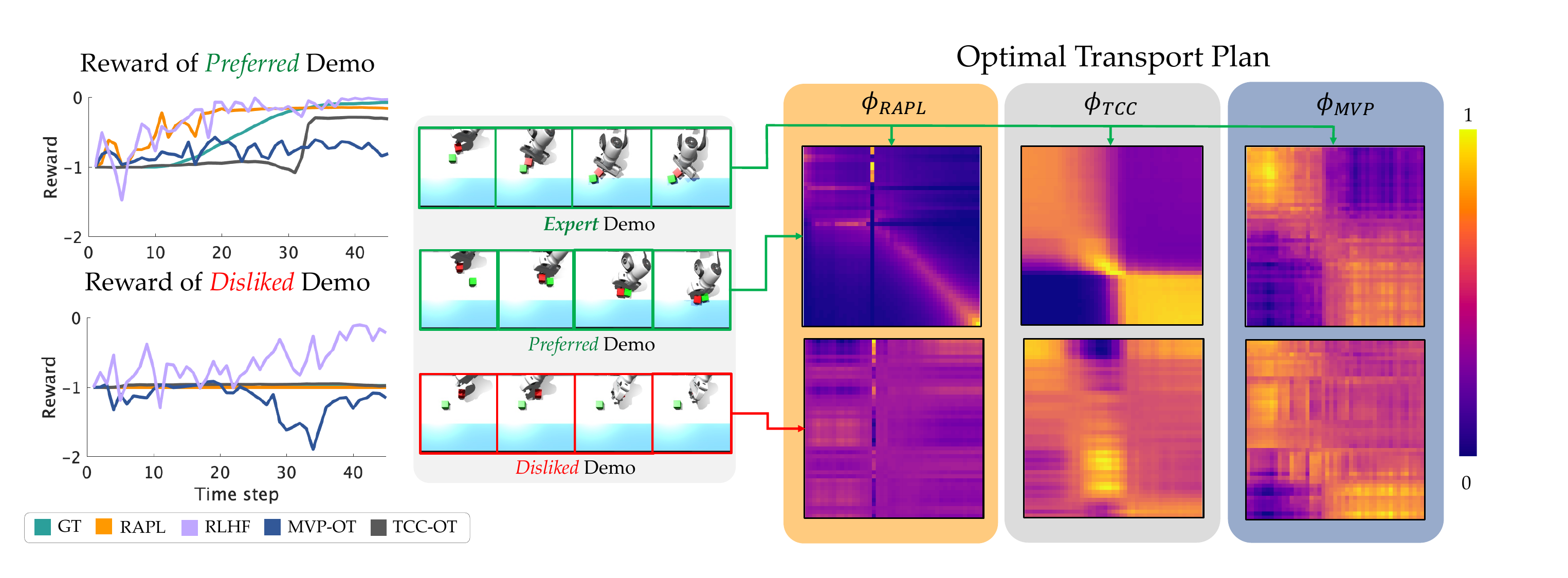}
    \caption{\textbf{Manipulation: Reward Prediction. }   
    (center) Expert, preferred, and disliked video demo.
    (left) Reward of each video under each method. RAPL's predicted reward follows the GT pattern. RLHF assigns high reward to disliked behavior.
    (right) OT coupling matrix for each visual representation. Columns are embedded frames of expert demo. Rows of top matrices are embedded frames of preferred demo; rows of bottom matrices are embedded frames of disliked demo. Peaks exactly along the diagonal indicate that the frames of the two videos are aligned in the latent space; uniform values in the matrix indicate that the two videos cannot be aligned (i.e., all frames are equally ``similar’’ to the next). RAPL exhibits the diagonal peaks for expert-and-preferred and uniform for expert-and-disliked, while baselines show diffused values no matter the videos being compared.}
    \label{fig:franka_OT_rew_indomain}
\end{figure*}



\begin{figure*}[h!]
    \centering
    \includegraphics[width=1\textwidth]{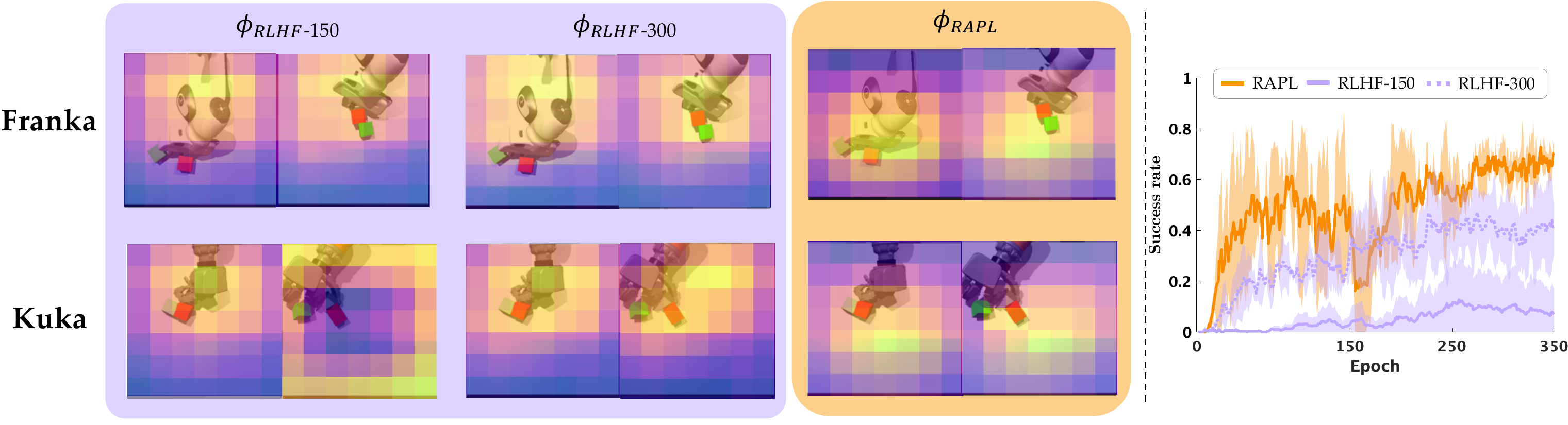}
    \caption{(Left) \textbf{Manipulation: Qualitative RLHF Comparison.} We visualize the attention map for RLHF-150 demos, RLHF-300 demos, and RAPL with 150 demos for both Franka and Kuka (cross-embodiment). Each entry shows two observations from respective demonstration set with the attention map overlaid. Bright yellow areas indicate image patches that contribute most to the final embedding; darker purple patches indicate less contribution. $\phi_{RLHF-150}$ is biased towards paying attention to irrelevant areas that can induce spurious correlations; in contrast RAPL learns to focus on the task-relevant objects and the goal region. $\phi_{RLHF-300}$'s attention is slightly shifted to objects but still pays high attention to the robot embodiment. (Right) \textbf{Manipulation: Quantitative RLHF Comparison.} \rapl outperforms \rlhf by $75\%$ with $50\%$ less human preference data.}
    \label{fig:attention_map}
\end{figure*}

\addspace
\noindent
\textbf{Results.} 
Figure~\ref{fig:xmagical_indomain_reward} shows the rewards over time for three example video observations in the \textbf{avoid} (left) and \textbf{group} task (right). Each video is marked as preferred by the end-user's ground-truth reward or disliked. 
Across all examples, \rapl's rewards are highly correlated with the \gt rewards: when the behavior in the video is disliked, then reward is low; when the behavior is preferred, then the reward is increasing. \tcc's reward is correlated with the robot making spatial progress (i.e., plot (E) and (F) where observations get closer to looking like the goal image), but it incorrectly predicts high reward when the robot makes spatial progress but violates the human's preference ((C) and (D) in \autoref{fig:xmagical_indomain_reward}). \rlhf performs comparably to \rapl, with slight suboptimality in scenarios (C) and (D). 
Figure~\ref{fig:xmagical_indomain_policy} shows the policy evaluation success rate during RL training with each reward function (\rb{solid line is the mean, shaded area is the standard deviation, over 5 trials with different random seeds.}). Across all environments, \rapl performs comparably to \gt (\textbf{avoid} success: $\approx80\%$, \textbf{group} success: $\approx60\%$) and significantly outperforms all baselines with better sample efficiency, RAPL takes 10 epochs to reach 70\% success rate in the \textbf{avoid} task (GT requires 100) and takes 100 epochs to reach 40\% success rate in the \textbf{avoid} task (GT requires 150), supporting our hypothesis.

\subsubsection{Robot Manipulation Experiments}
\label{subsec:in-domain-manipulation}

\begin{center}
\justifying
In the X-Magical toy environment, RAPL outperformed progress-based visual rewards, but direct preference-based reward prediction was a competitive baseline. Moving to the more realistic robot manipulation environment, we want to 1) disentangle the benefit of our fine-tuned representation from the optimal transport reward structure, and 
2) understand if our method still outperforms direct reward prediction in a more complex environment?
\end{center}

\noindent
\textbf{Task.} We design a robot manipulation task in the IsaacGym physics simulator \citep{makoviychuk2021isaac}. We replicate the tabletop \textbf{grouping} scenario, where a Franka robot arm needs to learn that the end-user prefers objects be pushed \textit{efficiently together} (instead of one-at-a-time) to the goal region (light blue region in the middle of \autoref{fig:x-magical-env}). 
In addition, we consider a more complex robot manipulation task---named \textbf{clutter} ---to further validate \rapl's ability to disentangle visual features that underlie an end-user's preferences. 
We increase the difficulty of the the \textbf{grouping} environment by adding visual distractors that are irrelevant to the human's preferences (left figure in Figure~\ref{fig:x-magical-env}). 
The environment has multiple objects on the table of various colors---red, green, and goal-region-blue---and some of the objects are cubes while others are rectangular prisms.
The Franka robot arm needs to learn that the end-user prefers to push the \textit{rectangular} objects (instead of the cubes) \textit{efficiently together} (instead of one-at-a-time) to the goal region.

\addspace
\noindent
\textbf{Privileged state \& reward.}
For \textbf{grouping}, the state $s$ is 18D: robot proprioception ($\theta_{\mathrm{joints}}\in \mathbb{R}^{10}$), 3D object positions
($p_{\mathrm{obj^{1,2}}}$), and object distances to goal ($d_{\mathrm{goal2obj^{1,2}}}$). The \textbf{grouping} reward is identical as in Section~\ref{subsec:in-domain-xmagical}.
For \textbf{clutter}, the state $s$ is 34D: robot proprioception ($\theta_{\mathrm{joints}}\in \mathbb{R}^{10}$), 3D object positions
($p_{\mathrm{obj_{rect}^{1,2}}}, p_{\mathrm{obj_{cube}^{1,\dots,4}}}$), and object distances to goal ($d_{\mathrm{goal2obj_{rect}^{1,2}}}, d_{\mathrm{goal2obj_{cube}^{1,\dots,4}}}$). The simulated human's reward is:
\begin{align*}
    &\rewh_{\mathrm{group}}(s) = - \max(d_{\mathrm{goal2obj^{rect,1}}}, d_{\mathrm{goal2obj^{rect,2}}}) \\  & ~~ - ||p_{\mathrm{obj_{rect}^{1}}} - p_{\mathrm{obj_{rect}^{2}}}||_2  - 0.1 \sum_{i=1}^{4} ||p_{\mathrm{obj_{cube}^{i}}} - p_{\mathrm{obj_{cube}^{i, init}}}||_2.
\end{align*}

\noindent
\textbf{Baselines.} In addition to comparing \rapl against (1) \gt and (2) \rlhf, we ablate the \textit{representation model} but control the \textit{visual reward structure}. We consider five additional baselines that all use optimal transport-based reward but operate on different representations: (3) \mvpot which learns image representation via masked visual pre-training; (4) \mvpotfinetune, which fine-tunes MVP representation  using images from the task environment; (5) \rtm, which is an off-the-shelf ResNet-18 encoder \citep{nair2022r3m} pre-trained on the Ego4D data set \citep{grauman2022ego4d} via a learning objective that combines time contrastive learning, video-language alignment, and a sparsity penalty; (6): \imagenet, which is a ResNet-18 encoder pre-trained on ImageNet; (7) \tccot \citep{dadashi2020primal} which embeds images via the TCC representation trained with 500 task demonstrations. We use the same preference dataset with $150$ triplets for training RLHF and RAPL.

\addspace
\noindent
\textbf{Visual model backbone.} All methods except MVP-OT and Fine-Tuned-MVP-OT share the same ResNet-18 visual backbone and have the same training setting as the one in the X-Magical experiment. MVP-OT and Fine-Tuned-MVP-OT use an off-the-shelf visual transformer \citep{xiao2022masked} pre-trained on the Ego4D data set \citep{grauman2022ego4d}. All representation models are frozen during policy learning.

\addspace
\noindent
\textbf{Hypotheses}. \textbf{H1:} \textit{\rapl's higher policy success rate are driven by its aligned visual representation}. \textbf{H2:} \textit{\rapl outperforms \rlhf with less human preference data}.

\addspace
\noindent
\textbf{Results: Reward Prediction.} In the center of \autoref{fig:franka_OT_rew_indomain} we show three video demos in the grouping task: an expert video demonstration, a preferred video, and a disliked video. 
On the right of Figure~\ref{fig:franka_OT_rew_indomain}, we visualize the optimal transport plan comparing the expert video to the disliked and preferred videos under three represenative visual representations, $\phi_{\rapl}$, $\phi_{\tccot}$, $\phi_{\mvpot}$.
Intuitively, peaks exactly along the diagonal indicate that the frames of the two videos are aligned in the latent space; uniform values in the matrix indicate that the two videos cannot be aligned (i.e., all frames are equally ``similar’’ to the next).
\rapl's representation induces precisely this structure: diagonal peaks when comparing two preferred videos and uniform when comparing a preferred and disliked video. 
Interestingly, we see diffused peak regions in all transport plans under both \tccot and \mvpot representations, indicating their representations struggle to align preferred behaviors and disentangle disliked behaviors.
This is substantiated by the left of \autoref{fig:franka_OT_rew_indomain}, which shows the learned reward over time of preferred video and disliked video. Across all examples, \rapl rewards are highly correlated to \gt rewards while baselines struggle to disambiguate.

We further conducted a quantitative analysis to investigate the relationship between the learned visual reward and the end-user’s ground-truth reward. For the robot manipulation tasks Franka Group and Franka Clutter, we computed the average Spearman’s correlation coefficient between the ground-truth reward trajectory and any other approach's reward trajectory across 100 video trajectories (showed in the left two columns of \autoref{table:Spearman}). We found that \rapl's learned visual reward shows the strongest correlation to the \gt reward compared to baselines.

\renewcommand{\arraystretch}{1.3}
\begin{table}[t!]
    \centering
    \Large
    \resizebox{0.49\textwidth}{!}{%
    \begin{tabular}{p{1.2in} | c  c | c  }
    \hline
       & \multicolumn{3}{|c}{ \textbf{Spearman's Correlation} $\uparrow$ } \\ \hline 
         & \textbf{Franka Group} & \textbf{Franka Clutter} & \textbf{Kuka Group}\\ \hline
        \rapl & \cellcolor[HTML]{c4f2cb}\textbf{0.59} &  \cellcolor[HTML]{c4f2cb}\textbf{0.61} & \cellcolor[HTML]{c4f2cb}\textbf{0.47}\\ \hline
        \rlhf & 0.38 &  0.26 & 0.31\\ \hline
        \mvpot &  -0.1&  0.08 & 0.02\\ \hline
        \mvpotfinetuneabbrev &  0.19&  0.11& 0.02 \\ \hline
        \imagenetabbrev &  -0.09 &  -0.02 & 0.12\\\hline
        \rtm & 0.03 &  -0.17 &  -0.14 \\ \hline
    \end{tabular}
    }
    \vspace{0.5em}
    \caption{\rb{Spearman's rank correlation coefficient between the \gt reward and each learned visual reward.}}
    \vspace{-0.8em}
    \label{table:Spearman}
\end{table}

\begin{figure}[h!]
\centering
\includegraphics[width=0.5\textwidth]{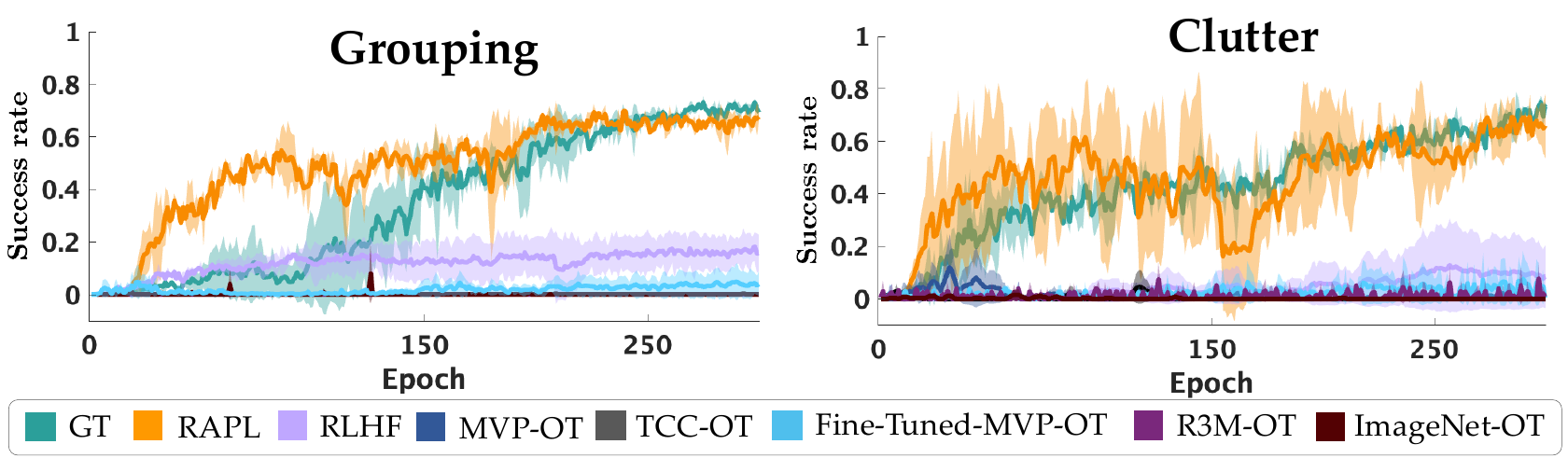}
\caption{\textbf{Manipulation: Policy Learning.} Success rate during robot policy learning under each visual reward. }
\vspace{-1em}
\label{fig:franka_success_v_reward}
\end{figure}

\addspace
\noindent
\textbf{Results: Policy Learning.} \autoref{fig:franka_success_v_reward} shows the policy evaluation history during RL training with each reward function. Across all the manipulation environments, we see \rapl performs comparably to \gt (succ. rate: $\approx 70\%)$ while all baselines struggle to achieve a success rate of more than $\approx 10\%$ with the same number of epochs, supporting \textbf{H1}.


\addspace
\noindent
\textbf{Results: Sample Complexity of \rapl vs. \rlhf.}
\begin{figure*}[t!]
    \centering
    \includegraphics[width=\textwidth]{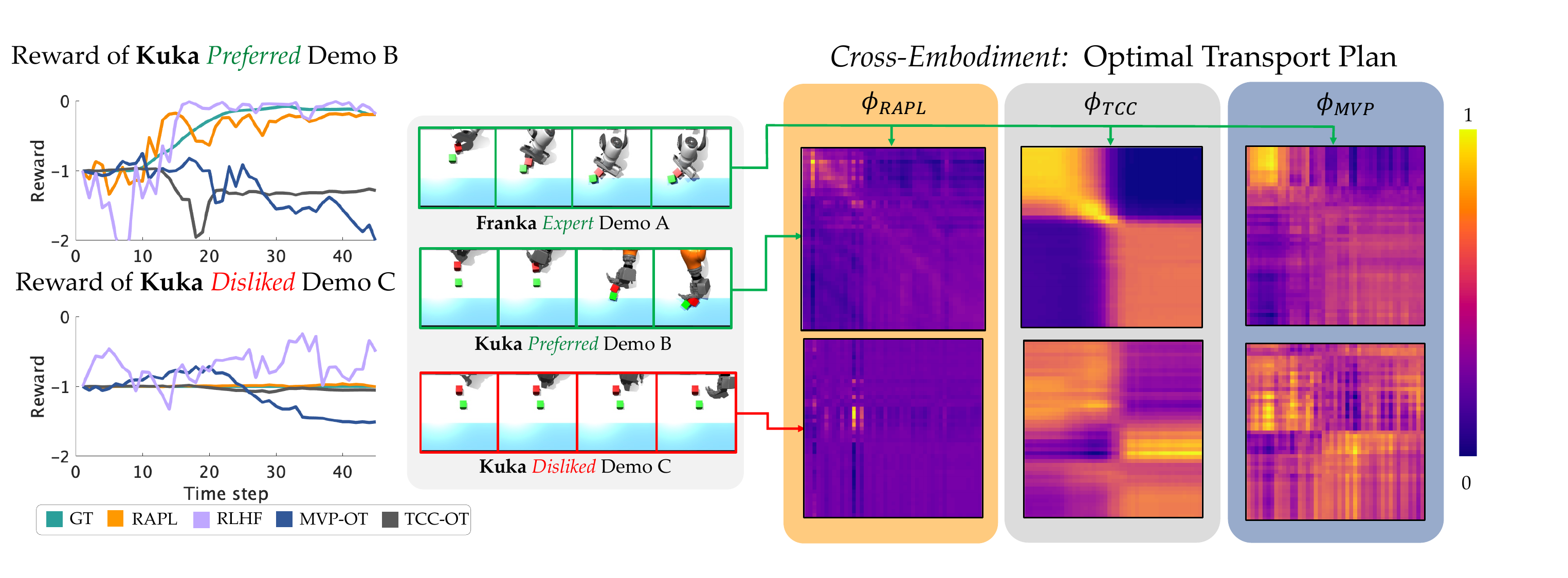}
    \vspace{-0.5cm}
    \caption{\textbf{Manipulation: Cross-Embodiment Reward Transfer.}
    (center) Expert video of \textit{Franka} robot, preferred video of \textit{Kuka}, and disliked \textit{Kuka} video. 
    (left) Predicted reward under each method trained only on \textit{Franka} video preferences. RAPL’s reward generalizes to the \textit{Kuka} robot and follows the GT pattern.
    (right) OT plan for each visual representation shown in the same style as in \autoref{fig:franka_OT_rew_indomain}. 
    RAPL's representation shows a diagonal OT plan for expert-and-preferred demos vs. a uniform for expert-and-disliked, while baselines show inconsistent plan patterns.}
    \label{fig:kuka_cross_domain_reward}
\end{figure*}

\begin{figure*}[t!]
    \centering
    \includegraphics[width=\textwidth]{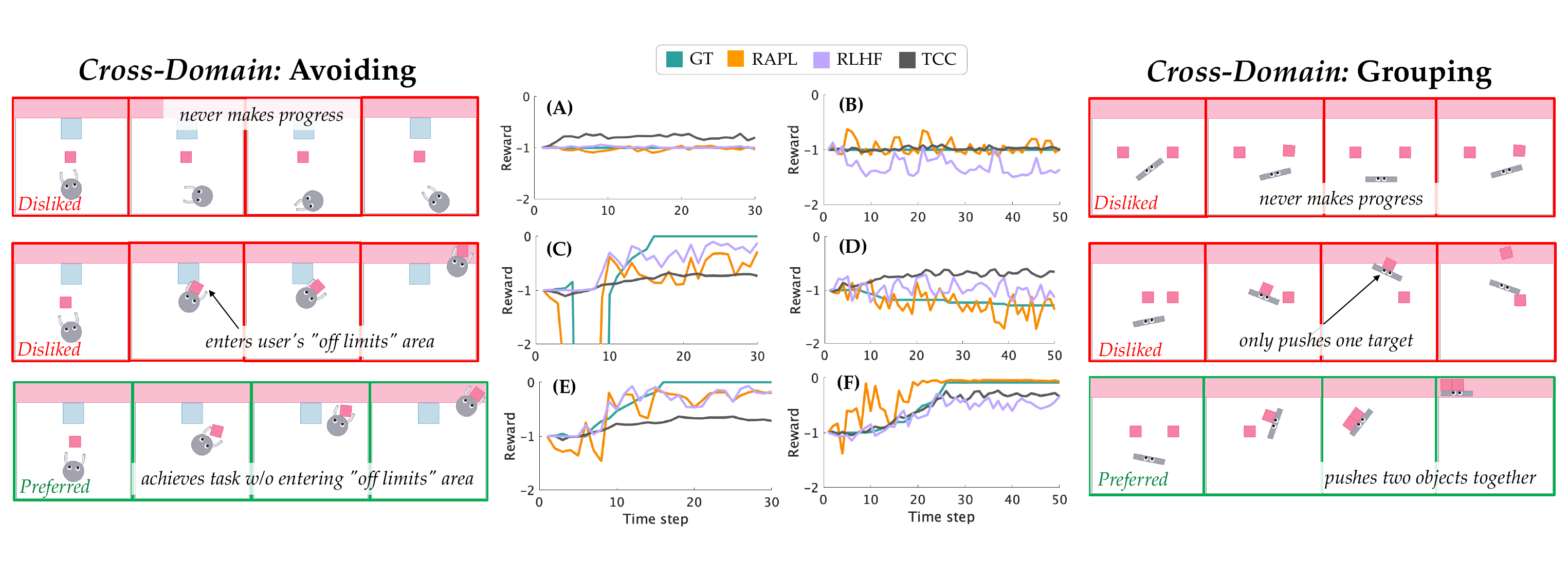}
    \vspace{-1cm}
    \caption{\textbf{X-Magical: Cross-Embodiment Reward Transfer.} \rapl discriminates preferred and disliked videos of novel robots. }
\label{fig:xmagical_crossdomain_reward}
\end{figure*}

It's surprising that \rlhf fails in a more realistic environment since its objective is similar to ours, but without explicitly considering representation alignment.
To further investigate this, we apply a linear probe on the final embedding and visualize the image heatmap of what \rapl's (our representation model trained with 150 training samples), \rlhf-150's (\rlhf trained with 150 samples), and \rlhf-300's (\rlhf trained with 300 samples samples) final embedding pays attention to in the top row of \autoref{fig:attention_map} (left). 
We see that $\phi_{RAPL}$ learns to focus on the objects, the contact region, and the goal region while paying less attention to the robot arm;
$\phi_{RLHF-150}$ is biased towards paying attention to irrelevant areas that can induce spurious correlations (such as the robot arm and background area);
$\phi_{RLHF-300}$'s attention is slightly shifted to objects while still pays high attention to the robot embodiment.
When deploying $\phi_{RLHF-300}$ in Franka manipulation policy learning (\autoref{fig:attention_map} (right)), we observe that policy performance is slightly improved, indicating that with more feedback data, preference-based reward prediction could yield an aligned policy. Nevertheless, \rapl outperforms \rlhf by $75\%$ with $50\%$ less training data, supporting \textbf{H2}.

While all the RAPL results above used 150 preference queries to train the representation, we also train a visual representation with 100, 50, and 25 preference queries.  
We measure the success rate of the robot manipulation policy trained for each ablation of RAPL. We find that RAPL-150 achieves $\approx 70\%$ success rate, RAPL-100 achieves $\approx 65\%$, RAPL-50 achieves $\approx 58\%$, and RAPL-25 achieves $\approx 45\%$ policy success rate despite using only 17\% of the original preference dataset.



\subsection{Results: Zero-Shot Generalization Across Robot Embodiments}
\label{sec:cross-domain-results}

So far, the preference feedback used to align the visual representation was given on videos $\obstraj \in \tilde{\Xi}$ generated on the same embodiment as that of the robot. 
However, in reality, the human could give preference feedback on videos of a \textit{different} embodiment than the specific robot's.
We investigate if our approach can \textit{generalize} to changes in the embodiment between the preference dataset $\tilde{S}_\human$ and the robot policy optimization. 

\begin{figure*}[h]
    \centering
    \vspace{-0.2cm}
    \includegraphics[width=1\textwidth]{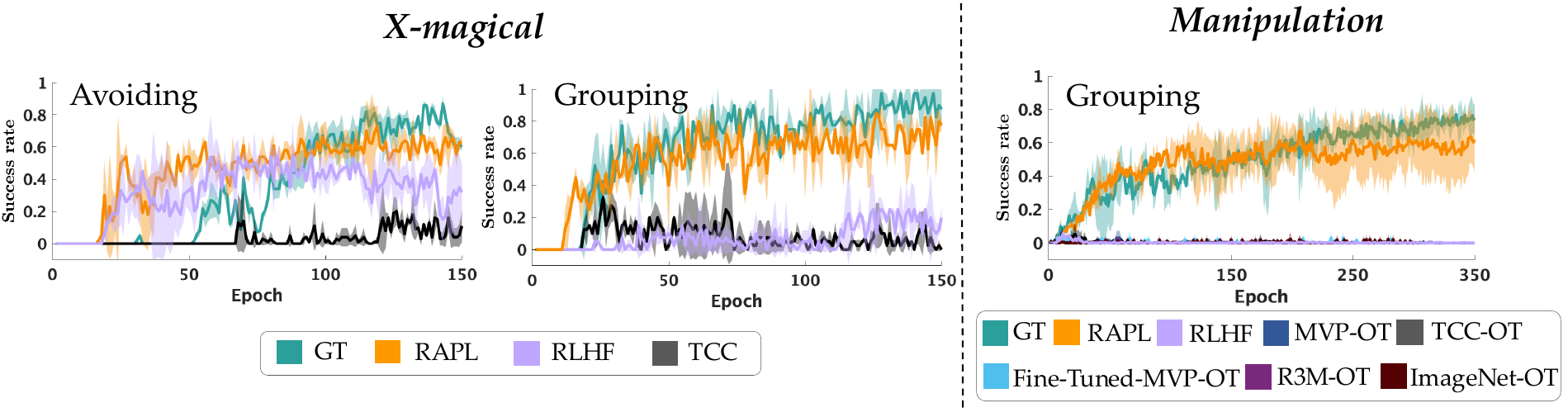}
    \vspace{-1.5em}
    \caption{\textbf{Cross-Embodiment Policy Learning.} Policy evaluation success rate during policy learning. Colored lines are the mean and variance of the evaluation success rate. RAPL achieves a comparable success rate compared to GT with high learning efficiency, and outperforms all baselines.}
    \vspace{-1em}
    \label{fig:xmagical_crossdomain_policy}
\end{figure*}

\addspace
\noindent
\textbf{Tasks \& baselines.} We use the same setup for each environment as in Section~\ref{sec:in-domain-results}.

\addspace
\noindent
\textbf{Cross-domain agents.} In X-Magical, reward functions are always trained on the \textit{short stick} agent, but the learning agent is a \textit{gripper} in \textbf{avoid} and a \textit{medium stick} agent in \textbf{grouping} task (illustrated in \autoref{fig:xmagical_crossdomain_reward}). In robot manipulation we train RAPL and RLHF on videos of the \textit{Franka} robot, but deploy the rewards on the \textit{Kuka} robot (illustrated in \autoref{fig:kuka_cross_domain_reward}).

\addspace
\noindent
\textbf{Hypothesis.} \textit{\rapl enables zero-shot cross-embodiment transfer of the visual reward compared to other baselines.}

\addspace
\noindent
\textbf{Results.}
\autoref{fig:xmagical_crossdomain_reward} and \autoref{fig:kuka_cross_domain_reward} show the learned rewards over time for the three cross-embodiment video observations (marked as preferred by the end-user’s ground-truth reward or disliked) in the cross-embodiment X-Magical environment and the manipulation environment. 
Across all examples, \rapl’s rewards are highly correlated with the \gt rewards even when deployed on a cross-embodiment robot. 
In the right-most column of \autoref{table:Spearman}, we show the average Spearman’s correlation coefficient between the
ground-truth reward trajectory and any other approach's reward trajectory across 100 videos in the cross-embodiment manipulation environment (grouping task). We found that \rapl's
learned visual reward shows the strongest correlation to the \gt reward compared to baselines.

In \autoref{fig:xmagical_crossdomain_policy}, we show the policy evaluation histories during RL training with each reward function in the cross-embodiment X-Magical environment and the manipulation environment. We see that in all cross-embodiment scenarios, \rapl achieves a comparable success rate compared to \gt and significantly outperforms baselines which struggle to achieve more than zero success rate, supporting our hypothesis. In the bottom row of \autoref{fig:attention_map}, we visualize the attention maps of \rapl and \rlhf. 
We see that $\phi_{RAPL}$ learns to focus on the objects, the contact region, and the goal region while paying less attention to the robot arm compared to \rlhf even when the encoder is learned from a Franka robot and deployed on a Kuka robot.


Furthermore, we note an interesting finding in the X-Magical grouping task when the representation is trained on videos of the \textit{short stick} agent, but the learning agent is the \textit{medium stick} agent (\autoref{fig:xmagical_crossdomain_reward}). 
Because the \textit{short stick} agent is so small, it has a harder time keeping the objects grouped together; in-domain results from Section~\ref{sec:in-domain-results} show a success rate of $\approx$ 60\% (see \autoref{fig:xmagical_indomain_policy}). 
In theory, with a well-specified reward, the task success rate should \textbf{increase} when the \textit{medium stick} agent does the task, since it is better suited to push objects together. 
Interestingly, when the \textit{short stick} visual representation is transferred zero-shot to the \textit{medium stick}, we see precisely this: \rapl's task success rate \textbf{improves} by 20\% under cross-embodiment transfer (succ. rate $\approx$ 80\% as shown in the middle plot of (\autoref{fig:xmagical_crossdomain_policy}). 
This indicates that RAPL can learn task-relevant features that can guide correct task execution even on a new embodiment.

\section{Aligning Diffusion Policies in the Real World with RAPL}
\label{sec:DPO-result}

In the previous section, we demonstrated how RAPL’s reward can align robot policies via reinforcement learning in simulation. 
However, high-fidelity simulators are often impractical for many real-world robotics tasks (e.g., deformable objects) and reinforcement learning in the real world is still an open research problem. 
Motivated by this, we turn to an algorithmic variant of RLHF, Direct Preference Optimization (DPO) described in Section \ref{sec:problem-setup}, for aligning visuomotor policies without a simulator. 
DPO updates the policy via contrastive learning using preference rankings on the behavior generations; however, to reliably update the policy it requires a significant amount of preference labels. 
In this section, we demonstrate how our RAPL reward enables the scalable generation of \textit{synthetic} preference rankings, significantly minimizing the number of real human labels while still achieving a high policy alignment.  


\begin{figure*}[t!]
    \centering
    \includegraphics[width=1\textwidth]{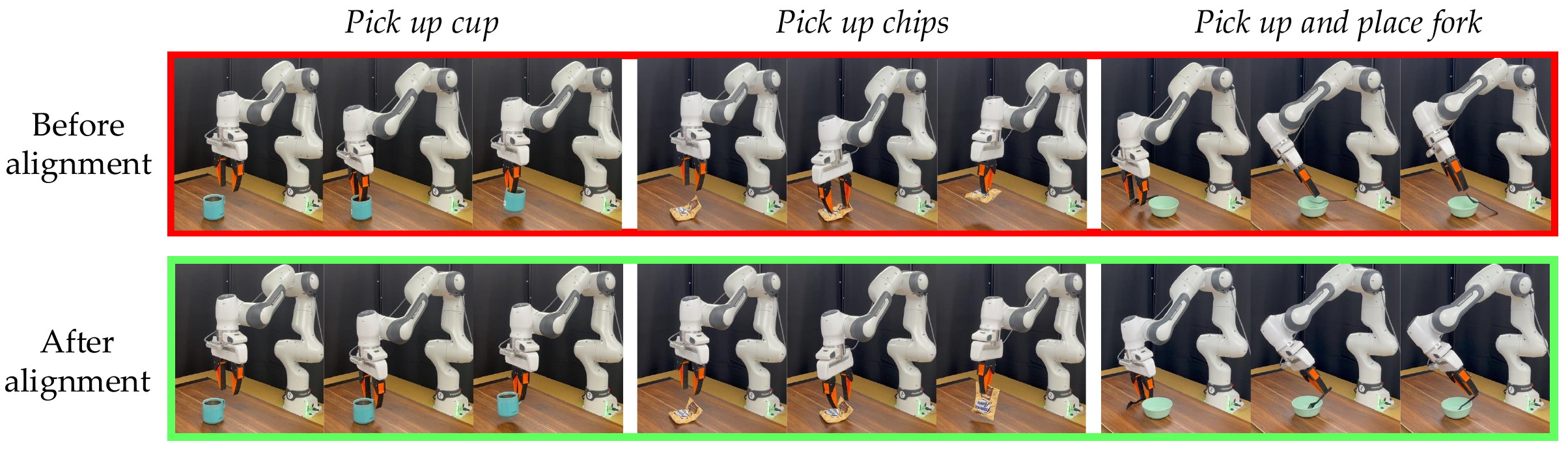}
    \vspace{-1.5em}
    \caption{\textbf{Diffusion Policy Alignment Results}. (Top) The pre-trained visuomotor policy exhibits undesired behaviors: grasping the interior of the cup (left), crushing the chips (middle), and making contact with the tines of the fork and dropping it out of the bowl (right). (Bottom) After alignment using RAPL rewards, the robot's behaviors are aligned with the end-user's preferences. }
    \vspace{-1em}
    \label{fig:DPO_behavior_visual}
\end{figure*}

\subsection{Experiment Design}

\textbf{Tasks}. We consider three real-world robot manipulation tasks. In the \textbf{picking up cup} task, the end-user prefers that the robot gripper utilizes the cup's handle to pick it up, avoiding contact with the interior of the cup to prevent contamination of the water inside. 
In the \textbf{picking up chips} task, the end-user prefers the robot gripper to hold the packaging by its edges rather than squeezing the middle, which may crush the chips.
In the \textbf{placing fork} task, the end-user prefers the robot gripper to pick the fork by the handle and gently place the fork into the bowl rather than picking by the tines or dropping it from an inappropriate height, which would cause the sanitation issue or forceful fall into the bowl. 

\addspace
\noindent
\textbf{Robot visuomotor policy}. We use a Diffusion Policy \citep{chi2024diffusionpolicy} as our robot visuomotor policy. The model takes the current and previous image observations from the wrist and third-person cameras as inputs to predict the distribution of a Franka robot's future motions. The image observations are encoded using a ResNet-19 visual encoder trained together with the policy network. 

\addspace
\noindent
\textbf{Initial policy training \& preference dataset construction}.
For each task, we collect 100 demonstrations covering multiple behavior modes to train the initial reference robot policy, which we denote as $\pi_{\text{ref}}$. 
After training, we rollout $\pi_{\text{ref}}$ under different initializations of the task by randomizing the task configurations (e.g., object states, robot states, etc.) to collect deployment video demonstrations. We query the end-user for \textbf{20 preference rankings} ($|\tilde{S}_{\reph}| = 20$) among the deployment video demonstrations to train the \rapl reward. We use our visual reward to automatically construct an order of magnitude more \textit{synthetic} preference rankings, $200$, for aligning the reference policy following \autoref{eq:dpo_update}.

\addspace
\noindent
\textbf{Independent \& dependent measures}. Throughout our experiments, we vary the visual reward signal used to construct preference rankings for policy alignment and the size of the preference dataset for representation learning.
To evaluate the robot's preference alignment in a task, we assess its behavior across $15$ random task configurations. For each configuration, the robot samples $100$ motion plans from its diffusion policy, aggregates them into $4$ modes using the non-maximum suppression (NMS) scheme from \cite{seff2023motionlm}, and executes the most likely mode. The end-user then grades the robot's behavior as good or bad, and we measure the frequency of which the executed mode aligns with the end-user's preference.

\addspace
\noindent
\textbf{Hypothesis.} \textit{\rapl generates accurate synthetic preference rankings from few real human preferences and enables better policy alignment  compared to baselines.}

\addspace
\noindent
\textbf{Baselines.} Similar to the evaluations in Section~\ref{sec:in-domain-results}, we consider the following visual reward baselines for constructing the preference rankings: (1) \gt (we manually label \textbf{100} preference rankings); (2) \rlhf (trained on the same \textbf{20} preference rankings that RAPL is trained on); (3) \rlhfl (RLHF but with a 3x larger preference dataset of \textbf{60} rankings); (4) \tccot; (5) \rtm; and (6) \mvpot.  


\subsection{Results}

\textbf{Results: Qualitative.}
\autoref{fig:DPO_behavior_visual} visualizes robot behaviors under both the initial visuomotor policy and the policy aligned using synthetic preference rankings constructed by \rapl across three manipulation tasks.
In the top row, we observe that the reference policy, while completing the tasks, exhibits undesired behaviors: contacting the interior of the cup (left), crushing the chips (middle), and making contact with the tines of the fork, resulting in the fork being dropped out of the bowl (right). After aligning the policy with the \rapl reward, the robot exhibits behaviors that align with the end-user's preferences: it picks up the cup by the handle, grasps the chips by the corner of the bag, and holds the handle of the fork when placing it in the bowl.

\medskip 
\noindent
\textbf{Results: Quantitative.} We present the alignment scores of the reference and each fine-tuned policy in \autoref{tab:alignment_score}. 
The results show that robot policies aligned using preference rankings constructed by \rapl achieve significantly higher alignment scores compared to all baselines across the three tasks and demonstrate performance comparable to \gt but with \textbf{5x} less human annotations. 
Notably, preference rankings labeled by the vanilla \rlhf show minimal improvement and, in some cases, degrade the alignment score of the initial reference policy. This suggests that \rlhf fails to learn accurate rewards capable of distinguishing between preferred and non-preferred behaviors. 
While \rlhfl does improve the reference policy's alignment, it uses $\textbf{3x}$ more preference rankings than \rapl to learn the visual reward, supporting our hypothesis that \rapl minimizes human feedback while maximizing alignment.

\renewcommand{\arraystretch}{1.7}
\begin{table}[h!]
\centering
\large
\resizebox{0.48\textwidth}{!}{
\begin{tabular}{l|c c c c c c c c c}
\toprule
 & & &\multicolumn{5}{c}{ \textbf{Behavior Alignment Score} ($\uparrow$) } \\ \hline 
  & \textbf{Ref.} & \gt & \rapl & \rlhf & \rlhfl & \tccot &  \rtm & \mvpot\\
\toprule
\textbf{Cup}  & \cellcolor[HTML]{d4d4d4}0.2 & \cellcolor[HTML]{d4d4d4}0.7 & \cellcolor[HTML]{c4f2cb}\textbf{0.8} & 0.3 & 0.7 & 0.4 & 0.4 & 0.5\\ 
\textbf{Fork}  & \cellcolor[HTML]{d4d4d4}0.2 & \cellcolor[HTML]{d4d4d4}0.7 & \cellcolor[HTML]{c4f2cb}\textbf{0.5} & 0.1 & 0.4 & 0.1 & 0.0 & 0.2\\
\textbf{Bag} & \cellcolor[HTML]{d4d4d4}0.4 & \cellcolor[HTML]{d4d4d4}0.6 & \cellcolor[HTML]{c4f2cb}\textbf{0.7} & 0.4 & 0.4 & 0.6 & 0.1 & 0.4\\
\toprule
\end{tabular}
}
\vspace{0.5em}
\caption{\textbf{Hardware Experiments: Behavior Alignment Score.} Robot visuomotor policies aligned using RAPL outperform all baselines and demonstrate performance comparable to GT but with \textbf{5x} less human annotations.}
\label{tab:alignment_score}
\end{table}


\section{Conclusion}

In this work, we presented Representation-Aligned Preference-based Learning (RAPL), an observation-only, human data-efficient method for aligning visuomotor policies. 
%
Unlike traditional RLHF, RAPL focuses human preference feedback on fine-tuning pre-trained vision encoders to align with the end-user's visual representation and then constructs a dense visual reward via feature matching in this aligned representation space. 
We validated the effectiveness of RAPL through extensive experiments in both simulated and real-world settings. 
In simulation, we showed that RAPL can  learn high-quality rewards with half the amount of human data traditional RLHF approaches use, and the learned rewards can generalize to new robot embodiments. 
In hardware experiments, we demonstrated that RAPL can successfully align pre-trained Diffusion Policies for three object manipulation tasks while needing \textbf{5x} less human preference data than prior methods.
We hope that our first steps with RAPL will bolster more research on how to align next-generation visuomotor policies with end-user needs while reducing human labeling burden. 

\bibliographystyle{SageH}
\bibliography{ref}


\end{document}